\pdfoutput=1

\documentclass[11pt]{article}

\usepackage[]{EMNLP2022}

\usepackage{times}
\usepackage{latexsym}

\usepackage[T1]{fontenc}

\usepackage[utf8]{inputenc}

\usepackage{microtype}

\usepackage{inconsolata}

\usepackage{graphicx}
\usepackage{subfigure}

\title{Code Comment Inconsistency Detection with BERT and Longformer}

\author{Theo Steiner \\
  UNC Chapel Hill \\
  \texttt{theo2023@live.unc.edu} \\\And
  Rui Zhang \\
  Penn State University \\
  \texttt{rmz5227@psu.edu} \\}

\begin{document}
\maketitle
\begin{abstract}
Comments, or natural language descriptions of source code, are standard practice among software developers. By communicating important aspects of the code such as functionality and usage, comments help with software project maintenance. However, when the code is modified without an accompanying correction to the comment, an inconsistency between the comment and code can arise, which opens up the possibility for developer confusion and bugs. In this paper, we propose two models based on BERT \citep{Devlin:19} and Longformer \citep{Beltagy:20} to detect such inconsistencies in a natural language inference (NLI) context. Through an evaluation on a previously established corpus of comment-method pairs both during and after code changes, we demonstrate that our models outperform multiple baselines and yield comparable results to the state-of-the-art models that exclude linguistic and lexical features. We further discuss ideas for future research in using pretrained language models for both inconsistency detection and automatic comment updating.
\end{abstract}

\section{Introduction}
\label{sec:introduction}
\begin{figure}[t]
    \centering
    \subfigure[Inconsistent]{\includegraphics[width=\linewidth]{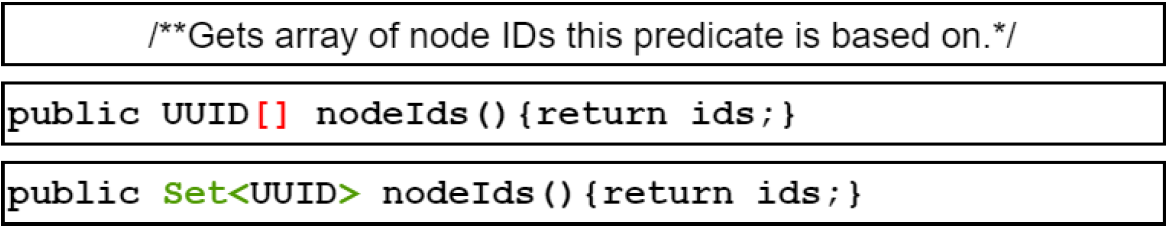}}
    \subfigure[Consistent]{\includegraphics[width=\linewidth]{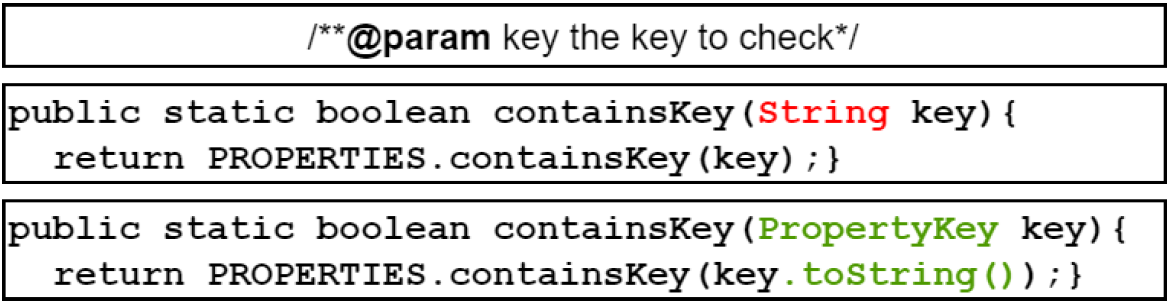}}
    \caption{Inconsistent and consistent Javadoc comments after each code method is modified \citep{Panthaplackel:21}. Removed tokens are in red and added tokens are in green.}
    \label{fig:comments}
\end{figure}

Natural language comments embedded within source code are indispensable to successful software projects. Although they are not compiled, comments serve as a guide to developers by providing valuable information about the functionality the source code implements as well as how developers are meant to utilize the code \citep{Tan:12}. Comments additionally help reduce ramping-up time, which is the time required for new developers to get accustomed to the specifics of a software project. Moreover, since developers on average spend around 82\% of their time on program comprehension and navigation \citep{Xia:18}, maintaining consistent comments can increase developer productivity. As such, it is essential for developers to keep comments up to date with the code so that their benefits can be capitalized on. In practice, however, comments are not always updated alongside code changes \citep{Wen:19}, which can lead to them becoming outdated and inconsistent with the code. Figure \ref{fig:comments} shows an example of an inconsistent comment versus a consistent comment. If not addressed in a timely manner, these inconsistencies can linger in code bases and mislead developers, becoming sources of confusion, bugs, and an overall impaired software development cycle (\citealp{Wen:19}; \citealp{Jiang:06}; \citealp{Tan:07}; \citealp{Ibrahim:12}). Code comment inconsistency detection is therefore of immense practical use to software developers who have a vested interest in keeping their code bases easily readable, navigable, and as bug-free as possible.

Prior work focuses on rule-based techniques for specific types of comments such as references to renamed identifiers \citep{Ratol:17}, API directives pertaining to parameter constraints \citep{Zhou:17}, and \verb|null| values and exceptions \citep{Tan:12}. Others have attempted text similarity methods (\citealp{Rabbi:20}; \citealp{Corazza:18}). However, these previous approaches only apply to pre-existing inconsistencies within a software project, which is not helpful for developers who would benefit more from a comprehensive, real-time comment maintenance system.

To resolve this dilemma, \citet{Panthaplackel:21} develop a state-of-the-art deep learning approach that not only detects existing inconsistencies (referred to as \emph{post hoc}) but also flags inconsistencies that appear immediately as a result of code changes, before they can cause harm to the code base (referred to as \emph{just-in-time}). A comment update model, designed to aid developers in fixing discrepancies by suggesting updates to those comments that are labeled as inconsistent, is also presented within an extrinsic evaluation \citep{Panthaplackel:21}.

In this paper, we alternatively propose a much more straightforward architecture and methodology. Our approach leverages existing large, pretrained models, namely BERT \citep{Devlin:19} and Longformer \citep{Beltagy:20} to address the problem of code comment inconsistency detection as a natural language inference (NLI) task.\footnote{Implementation is available at \url{https://github.com/theo2023/coco-bert-longformer}.} It is well known that such models have previously demonstrated high performance on many downstream natural language processing (NLP) tasks, including NLI. To compare with the state-of-the-art, we evaluate our models on the corpus provided by \citet{Panthaplackel:21} and show that in the post hoc setting, our approach outperforms the baselines by substantial margins but underperforms the state-of-the-art in the just-in-time setting.

Our main contributions are that we (1) formulate code comment inconsistency detection as an NLI problem via binary classification; (2) show that large, pretrained language models can provide significant improvements in performance in the post hoc setting and comparable results to the state-of-the-art (without the addition of explicit features) in the just-in-time setting; and (3) offer proposals for future work in inconsistency detection and comment updating based on our approach.

\section{Related Work}
\label{sec:relatedwork}
\citet{Rabbi:20} introduce a siamese recurrent neural network architecture that detects inconsistencies by measuring the similarity between the comment and the code. They employ two separate LSTM \citep{Hochreiter:97} models, one for the comment and the other for the code, and compute the Euclidean norm distance between the final hidden vectors of each LSTM. If the similarity score is below a certain predefined threshold, the comment is classified as inconsistent. While this technique achieves high recall on a limited set of Java code-comment pairs, it does not consider changes between different versions of the code and thus is incompatible with the just-in-time setting. Additionally, in general, text similarity is only a relevant metric for comments that summarize their code methods, not comments that document more specific aspects of the code (\emph{e.g.}, \verb|@return| and \verb|@param| Javadoc comments) \citep{Panthaplackel:21}.

\citet{Liu:18} develop several random forest classifiers and use 64 features relating to the code before and after modification, the comment, and their relationship. The features include code refactorings, length of the comment, and code-comment similarity, among others, in order to detect inconsistent block and line comments just-in-time. We avoid such extensive feature engineering in our procedure and also consider Javadoc comments instead of the more low-level block/line comments.

Instead of a model based on code-comment similiarity or thorough feature extraction, \citet{Panthaplackel:21} learn the relationship between comments and code modifications through a more intricate architecture that features several BiGRU \citep{Cho:14} encoders and multi-head attention \citep{Vaswani:17} layers. This architecture is explained in more detail in Section \ref{sec:baselines}. Rather than using external attention modules, we rely on the built-in attention mechanisms of BERT and Longformer to correlate the comment and code tokens.

Automatic comment updating has also been of interest in the recent literature. This is the more challenging problem of resolving inconsistencies once they have been detected. Both \citet{Panthaplackel:20b} and \citet{Zhu:22} develop an encoder-decoder architecture that examines changes in the source code and generates an edit action sequence specifying how the comment should be updated, which is then parsed into the gold comment. Although we focus on inconsistency detection and do not provide a comment update model in this work, we encourage further research in this important area in Section \ref{sec:futurework}.

\section{Task}
The task of code comment inconsistency detection is presented as follows: given a comment $C$ and its corresponding code method $M$, determine whether $C$ is inconsistent with $M$. We accomplish this task in two settings, post hoc and just-in-time, and formulate it in terms of NLI.

\subsection{Post hoc}
In the post hoc setting, only the existing version of the comment-method pair is available such that inconsistencies are already present in the code base. Modifications to the code are not considered.

\subsection{Just-in-time}
\begin{figure}[t]
    \centering
    \includegraphics[width=\linewidth]{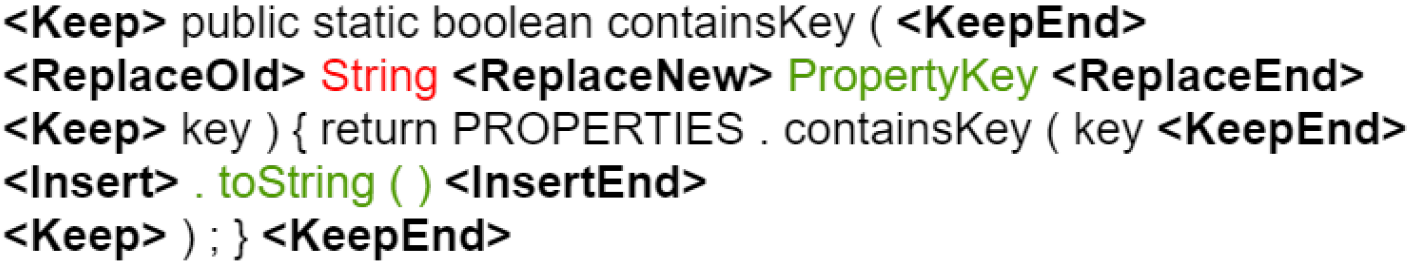}
    \caption{Sequence-based representation of $M_{edit}$ \citep{Panthaplackel:21}. Removed tokens are in red and added tokens are in green.}
    \label{fig:sequence}
\end{figure}

In the just-in-time setting, the idea is that modifications to the code must be analyzed in order to determine which code changes caused the comment to become inconsistent, and to detect this inconsistency \emph{before} it materializes and is committed to the repository. Thus, both versions of the code method $M_{old}$ and $M$ (before and after modification by the developer) are available, as well as $M_{edit}$, an edit action sequence representing the changes between $M_{old}$ and $M$. An example of $M_{edit}$ is shown in Figure \ref{fig:sequence}.

\subsection{Natural Language Inference}

We view code comment inconsistency detection as an NLI task, also known as textual entailment. Given a premise and hypothesis, the objective of this common NLP task is to determine whether the hypothesis entails the premise, contradicts the premise, or is neutral.\footnote{\url{https://paperswithcode.com/task/natural-language-inference}} For this specific problem, the premise is the comment $C$ and the hypothesis is the code method $M$. If our models predict that $M$ entails $C$, then $C$ is labeled as consistent. Otherwise, if our models predict that $M$ contradicts $C$, then $C$ is labeled as inconsistent. Note that we slightly modify NLI and drop the neutral label entirely, simplifying inconsistency detection to binary classification. Also note that in our initial problem formulation, the premise was $M$ and the hypothesis was $C$, but for practical reasons described in Section \ref{sec:ourmodels}, we swap the order of $M$ and $C$.

\section{Data}

We use the training, validation, and test data curated by \citet{Panthaplackel:21}.\footnote{\url{https://github.com/panthap2/deep-jit-inconsistency-detection}} This data consists of \verb|@return|, \verb|@param|, and summary Javadoc comments paired with their corresponding Java methods, totaling 40,688 examples (see Table \ref{datastats}). They consider comment-method pairs $(C_1,M_1),(C_2,M_2)$ from consecutive commits of popular open-source Java projects and set $C=C_1$, $M_{old}=M_1$, and $M=M_2$. Only examples in which $M_1\neq M_2$, \emph{i.e.}, where the developer modified the code method, are extracted. The data further includes a sequence of span diff subtokens for each method that we use as $M_{edit}$ in the just-in-time setting.

\begin{table}[t]
\setlength\tabcolsep{4.5pt}
\begin{tabular}{lcccc}
\hline
 & \textbf{Train} & \textbf{Valid} & \textbf{Test} & \textbf{Total} \\
 \hline
 \verb|@return| & 15,950 & 1,790 & 1,840 & 19,580 \\
 \verb|@param| & 8,640 & 932 & 1,038 & 10,610 \\
 Summary & 8,398 & 1,034 & 1,066 & 10,498 \\
 \hline
 Full & 32,988 & 3,756 & 3,944 & 40,688 \\
 \hline
 Projects & 829 & 332 & 357 & 1,518 \\
 \hline
\end{tabular}
\caption{\label{datastats}
Statistics of the full dataset \citep{Panthaplackel:21}.
}
\end{table}

Collected examples are given a positive label, \emph{i.e.}, $C_1$ is inconsistent with $M_2$, if $C_1\neq C_2$. The assumption behind this decision is that the developer updated the comment when they modified the code because the comment would have become inconsistent otherwise. On the other hand, collected examples are given a negative label, \emph{i.e.}, $C_1$ is consistent with $M_2$, if $C_1=C_2$. The assumption behind this decision is that the developer did not need to update the comment when they modified the code because the comment remained consistent \citep{Panthaplackel:21}.

However, this data collection procedure introduces noise in the form of mislabeled examples \citep{Panthaplackel:21}. A positive mislabeling occurs if the developer simply improved the comment without changing its semantics; hence, $C_1\neq C_2$ even though the comment is actually consistent. A negative mislabeling occurs if the developer neglected to update the comment alongside the code changes; hence, $C_1=C_2$ even though the comment is actually inconsistent. To reduce this noise, \citet{Panthaplackel:21} only consider the 1,000 best maintained Java projects and curate a smaller clean test sample separate from the full test set. As these clean examples were not provided, however, we do not report results from that data and only evaluate on the full test set.

Lastly, comment, method, and edit action sequences are tokenized via the subword tokenization algorithms of BERT and Longformer, \emph{i.e.}, WordPiece \citep{Schuster:12} and byte pair encoding \citep{Sennrich:16} respectively.

\section{Models}
We first discuss the baselines then introduce our models.

\subsection{Baselines} \label{sec:baselines}
\begin{itemize}
    \item \textbf{\citet{Corazza:18}}: This is a post hoc SVM approach with a TF-IDF schema that learns the decision boundary between consistent and inconsistent comment-method pairs.
    
    \item \textbf{CodeBERT BOW}: As implemented by \citet{Panthaplackel:21}, this bag-of-words baseline is the same underlying model for each setting. It is an application of CodeBERT \citep{Feng:20} in which the average embedding vectors of $C$ and $M$ (or $M_{edit}$) are concatenated and sent through a feedforward neural network.
    
    \item \textbf{\citet{Liu:18}}: Originally involving block and line comments that accompany code snippets, this is a just-in-time approach re-implemented by \citet{Panthaplackel:21}. It uses random forest classifiers with code, comment, and relationship features to predict inconsistent comments based on code changes.
    
    \item \textbf{\citet{Panthaplackel:21}}: There are nine total models. In the architecture, the code encoder (Figure \ref{fig:architecture} (3)) is one of two distinct configurations. The first is a sequence encoder, a BiGRU that encodes $M_{edit}$ (or $M$ in the post hoc setting). The $\textsc{Seq}(C,*)$ models correspond to the use of this encoder.

\begin{figure}[t]
    \centering
    \includegraphics[width=\linewidth]{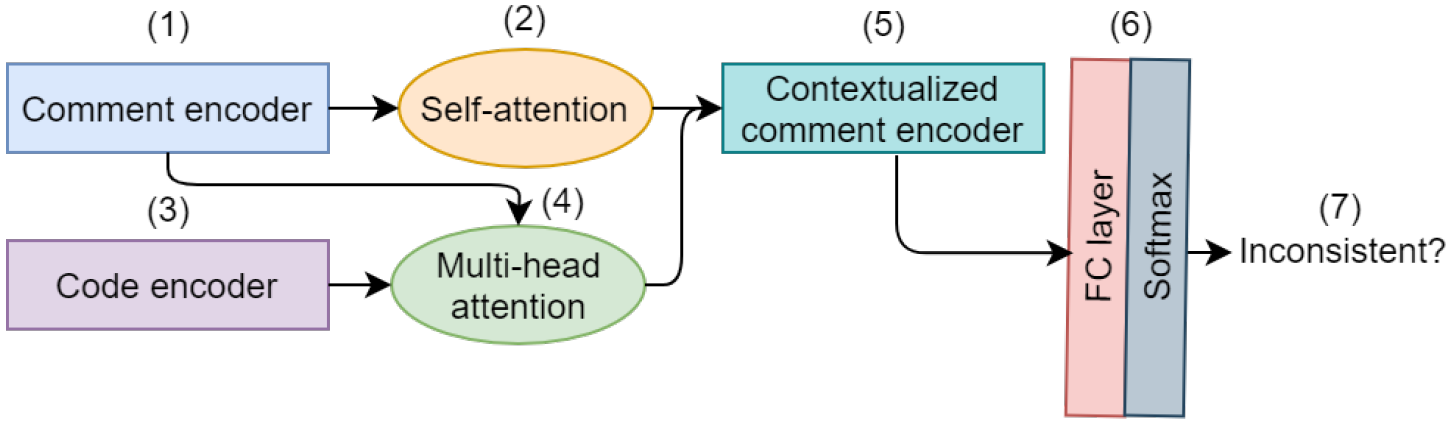}
    \caption{\citet{Panthaplackel:21} architecture.}
    \label{fig:architecture}
\end{figure}

\begin{figure}[t]
    \centering
    \includegraphics[width=\linewidth]{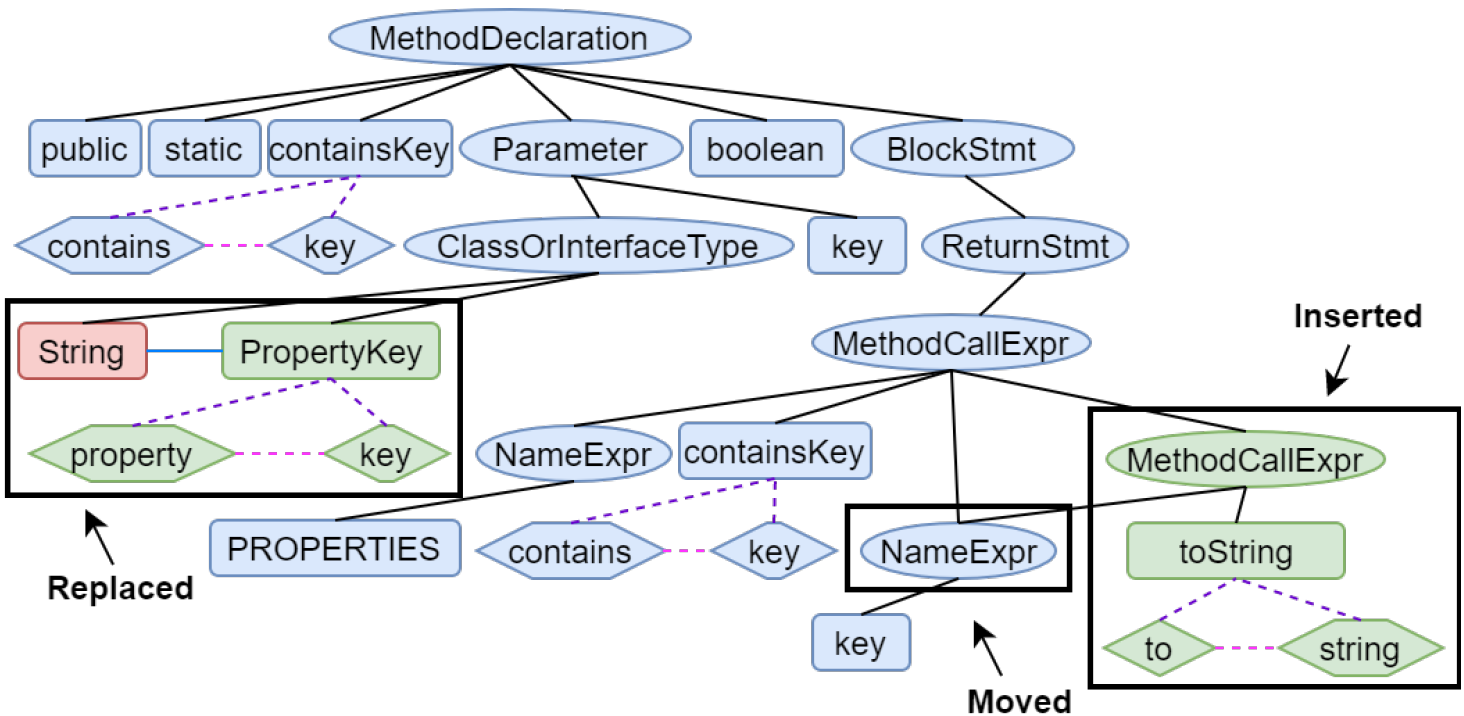}
    \caption{AST-based representation of $M_{edit}$ ($T_{edit}$) \citep{Panthaplackel:21}. Removed nodes are in red and added nodes are in green.}
    \label{fig:ast}
\end{figure}

    The second configuration is an abstract syntax tree (AST) encoder, a gated graph neural network (GGNN) \citep{Li:16} that encodes $T_{edit}$ (or $T$ in the post hoc setting), which is an AST representation of $M_{edit}$. Figures \ref{fig:sequence} and \ref{fig:ast} show examples of $M_{edit}$ and $T_{edit}$ respectively. The $\textsc{Graph}(C,*)$ models correspond to the use of this encoder. The last three models, denoted $\textsc{Hybrid}(C,*,*)$, incorporate both the GRU and GGNN when encoding the code method.
    
    Once the code and comment encoders encode their input as a sequence of tokens, integrating bidirectional context of other input tokens in the process, multi-head self-attention (Figure \ref{fig:architecture} (2)) and multi-head cross-attention (Figure \ref{fig:architecture} (4)) are each computed. The former updates the comment representation to include more context and help capture long-range dependencies, while the latter is employed between each hidden state of the comment encoder and the hidden states of the code encoder to associate the comment with changes in the code. Then, the contextualized embeddings from each attention module are combined and encoded by another BiGRU (Figure \ref{fig:architecture} (5)), at which point the comment encoding carries context from both the code method and other tokens in the comment. Finally, this output is fed through a fully connected layer and a softmax operation (Figure \ref{fig:architecture} (6)) to predict whether the comment is inconsistent with the code method (Figure \ref{fig:architecture} (7)).
\end{itemize}

\subsection{Our Models} \label{sec:ourmodels}
Our first model is BERT \citep{Devlin:19} finetuned on an NLI task. In other words, it is the base BERT model with an attached sequence classification head. Since BERT allows a maximum sequence length of 512 tokens, the concatenated sequences are truncated if they exceed that length. This technique is not ideal because potentially valuable information from tokens in $M$ (or $M_{edit}$) that the classifier may need to make a correct prediction is lost.

To better accommodate longer sequences, we propose our second model, Longformer \citep{Beltagy:20} finetuned on an NLI task. In other words, it is the base Longformer model with an attached sequence classification head. The eightfold increase in maximum sequence length (from 512 to 4096) provided by Longformer helps to reduce the adverse effects of truncation that our BERT model encounters. However, we believe the ability to process every single comment-method pair in the dataset does not justify the time and memory costs that would be incurred if we had chosen to leverage the default Longformer maximum sequence length of 4096. Therefore, the statistics in Table \ref{length98thpercentile} motivate a choice of 1024 as the custom maximum sequence length for our Longformer model. As in the first model, the concatenated sequences are truncated if they exceed this length, and lower performance is the tradeoff.

The major advantage of Longformer is its linear attention pattern, addressing the fact that the time and memory complexity of vanilla self-attention \citep{Vaswani:17} scales quadratically with sequence length \citep{Beltagy:20}. This improved attention mechanism makes Longformer especially efficient for longer sequences.

\begin{table}[t]
\setlength\tabcolsep{4.5pt}
\begin{tabular}{lcccc}
\hline
 & \verb|@return| & \verb|@param| & Summary & Full \\
 \hline
 $C$ & 34.0 & 32.0 & 35.0 & 34.0 \\
 $M_{old}$ & 728.0 & 936.8 & 753.1 & 797.0 \\
 $M$ & 726.0 & 942.8 & 750.1 & 790.0 \\
 $M_{edit}$ & 895.1 & 1102.6 & 955.3 & 981.0 \\
 \hline
\end{tabular}
\caption{\label{length98thpercentile}
98th percentile of the comment and code sequence lengths.
}
\end{table}

\begin{table}[t]
\setlength\tabcolsep{4.5pt}
\begin{tabular}{lcccc}
\hline
 & \verb|@return| & \verb|@param| & Summary & Full \\
\hline
$C$ & 9.7 & 8.4 & 13.3 & 10.3 \\
$M_{old}$ & 131.1 & 186.9 & 137.0 & 147.2 \\
$M$ & 131.9 & 187.7 & 135.4 & 147.3 \\
$M_{edit}$ & 179.4 & 240.9 & 186.6 & 197.3 \\
\hline
\end{tabular}
\caption{\label{avglengths}
Average lengths of the comment and code sequences \citep{Panthaplackel:21}.
}
\end{table}

As input to our models, the order $(C,M)$ was chosen because preliminary experiments revealed that $(M,C)$ did not lead to a significant change in results. We also found that with the $(M,C)$ ordering, there were cases in which all the tokens in $C$ were being truncated. Swapping the order of the inputs aligns with the fact that comments are shorter than their corresponding code methods on average (see Table \ref{avglengths}), so in the worst case, the end of $M$ is truncated rather than the entirety of $C$.

\section{Experiments}
\begin{table*}[t]
\centering
\begin{tabular}{lcccc}
\hline
\multicolumn{1}{c}{\textbf{Model}} & \textbf{Precision} & \textbf{Recall} & \textbf{F1} & \textbf{Accuracy}\\
\hline
\citet{Corazza:18} & 63.7 & 47.8 & 54.6 & 60.3 \\
CodeBERT BOW$(C,M)$ & 68.9 & 73.2 & 70.7 & 69.8 \\
$\textsc{Seq}(C,M)$ & 60.6 & 73.4 & 66.3 & 62.8 \\
$\textsc{Graph}(C,T)$ & 62.6 & 72.6 & 67.2 & 64.6 \\
$\textsc{Hybrid}(C,M,T)$ & 56.3 & 80.8 & 66.3 & 58.9 \\
\hline
$\textsc{Bert}(C,M)$ & 72.1 & 71.9 & 72.0 & 72.1 \\
$\textsc{Longformer}(C,M)$ & \textbf{92.7} & \textbf{81.0} & \textbf{86.4} & \textbf{87.3} \\
\hline
\end{tabular}
\caption{\label{posthoc}
Results for post hoc models compared with baselines. Scores are averaged across three different seeds.}
\end{table*}

\begin{table*}[t]
\centering
\begin{tabular}{lcccc}
\hline
\multicolumn{1}{c}{\textbf{Model}} & \textbf{Precision} & \textbf{Recall} & \textbf{F1} & \textbf{Accuracy}\\
\hline
CodeBERT BOW$(C,M_{edit})$ & 67.4 & 76.8 & 71.6 & 69.6 \\
\citet{Liu:18} & 77.5 & 63.8 & 70.0 & 72.6 \\
$\textsc{Seq}(C,M_{edit})$ & 80.7 & 73.8 & 77.1 & 78.0 \\
$\textsc{Graph}(C,T_{edit})$ & 79.8 & 74.4 & 76.9 & 77.6 \\
$\textsc{Hybrid}(C,M_{edit},T_{edit})$ & 80.9 & 74.7 & 77.7 & 78.5 \\
$\textsc{Seq}(C,M_{edit})$ + features & 88.4 & 73.2 & 80.0 & \textbf{81.8} \\
$\textsc{Graph}(C,T_{edit})$ + features & 83.8 & 78.3 & \textbf{80.9} & 81.5 \\
$\textsc{Hybrid}(C,M_{edit},T_{edit})$ + features & \textbf{88.6} & 72.4 & 79.6 & 81.5 \\
\hline
$\textsc{Bert}(C,M_{edit})$ & 76.4 & \textbf{78.3} & 77.3 & 77.1 \\
$\textsc{Longformer}(C,M_{edit})$ & 80.3 & 76.5 & 78.3 & 78.8 \\
\hline
\end{tabular}
\caption{\label{just-in-time}
Results for just-in-time models compared with baselines. Scores are averaged across three different seeds.
}
\end{table*}

\subsection{Implementation Details}

$C$ and $M$ (or $M_{edit}$) are tokenized separately and concatenated into one sequence of the form
\smallbreak
\centerline{[CLS] $C$ tokens [SEP] $M$ tokens [SEP],}
\smallbreak
where [CLS] is the classification token and [SEP] is the separation token. An attention mask is created as well as a binary mask representing the token type IDs so that BERT can differentiate between the part of the sequence that corresponds to $C$ and the part of the sequence that corresponds to $M$ (or $M_{edit}$). For the Longformer model, the token type IDs are unnecessary due to the use of the RoBERTa \citep{Liu:19} tokenizer. As described in Section \ref{sec:ourmodels}, the full sequence is truncated if it is longer than the maximum length permitted by the model. All other configurations, such as pretrained comment, code, and code edit embeddings, are the BERT and Longformer defaults.

\subsection{Training}

Because we are essentially training a binary classifier, we employ binary cross entropy (BCE) \citep{Zhang:18} as our loss function. We also use the Adam optimizer \citep{Kingma:14} and consider $\{$0.0001, 0.00001$\}$ as learning rates. To address memory issues, gradient accumulation is employed to achieve effective batch sizes of $\{$16, 32, 64$\}$. Training terminates if the validation F1 score does not improve for 10 epochs.

\subsection{Software and Hardware}

We implemented all models using PyTorch and the Hugging Face Transformers library,\footnote{\url{https://huggingface.co/docs/transformers/index}} and we used scikit-learn to compute the evaluation metrics precision, recall, F1 score, and accuracy. Models were trained in a distributed fashion on 8 NVIDIA RTX A5000 GPUs (24 GB each).

\section{Results and Analysis}

We report precision, recall, F1, and accuracy for our post hoc and just-in-time models in Tables \ref{posthoc} and \ref{just-in-time} respectively. In the post hoc setting, we find that both models outperform the baselines with respect to precision, F1, and accuracy, and $\textsc{Longformer}(C,M)$ achieves higher scores across all metrics. In the just-in-time setting, we find that the recall achieved by $\textsc{Bert}(C,M_{edit})$ of 78.3 matches the highest (from the baseline $\textsc{Graph}(C,T_{edit})$ + features); otherwise, our just-in-time models overall underperform the \citet{Panthaplackel:21} models that have added surface features like part-of-speech tags, comment/code overlap, etc. However, these models produce results comparable to the \citet{Panthaplackel:21} models without these features.

The results in Table \ref{posthoc} indicate that BERT and Longformer are clearly advantageous for post hoc inconsistency detection when finetuned on an NLI task. The substantial performance gains attained by our Longformer model are foreseeable given its linear attention pattern and ability to adequately process longer sequences. However, the decrease in performance from $\textsc{Longformer}(C,M)$ to $\textsc{Longformer}(C,M_{edit})$ in the just-in-time setting (Table \ref{just-in-time}) is puzzling, especially considering the increase in performance from $\textsc{Bert}(C,M)$ to $\textsc{Bert}(C,M_{edit})$. The reason for this performance drop is unclear.

\section{Future Work}
\label{sec:futurework}
Provided these results, we leave it to future work to both improve code comment inconsistency detection as an NLI task and apply our approach to a comment update model as mentioned in Sections \ref{sec:introduction} and \ref{sec:relatedwork}. Accordingly, in addition to the just-in-time system flagging the inconsistent comment and notifying the developer that they should manually correct it, the high-level overview is that the code changes that caused the inconsistency would be examined and suggestions would be generated to help the developer rectify the outdated comment.

Certain variants of these pretrained models also show promise. For instance, GraphCodeBERT \citep{Guo:21} is specifically designed for processing programming language and is pretrained on a large corpus of comment-code pairs. The main contribution of GraphCodeBERT is that it takes into account the semantic structure of code through its use of data flow. Furthermore, Longformer Encoder-Decoder \citep{Beltagy:20} is a possibility for a comment update model since it is intended for sequence-to-sequence tasks with longer input.

Another option as an implementation detail is to handle longer sequences via chunking, \emph{i.e.}, splitting the input sequence into smaller chunks and processing them separately. We did not pursue this method due to time constraints, but we believe it is worthwhile to explore.

\section{Conclusion}
\label{sec:conclusion}
We formulated code comment inconsistency detection as a natural language inference task and proposed two models based on BERT and Longformer. By evaluating on a known corpus of comment-method pairs in both a post hoc and just-in-time manner, we demonstrated that our models outperform several baselines in the post hoc setting and yield comparable results to the state-of-the-art in the just-in-time setting without additional linguistic and lexical features. We further discussed some ideas for future research in using pretrained language models for inconsistency detection and an automatic comment updating system.

\section*{Acknowledgements}
We thank the 2022 Machine Learning in Cybersecurity Research Experience for Undergraduates (REU) program at Penn State University, funded by National Science Foundation Grant 1950491, for supporting this work.

\bibliography{anthology,custom}
\bibliographystyle{acl_natbib}

\end{document}